# Fast Value Iteration for Goal-Directed Markov Decision Processes

Nevin L. Zhang and Weihong Zhang
Department of Computer Science
Hong Kong University of Science & Technology
{lzhang, wzhang}@cs.ust.hk

## Abstract

Planning problems where effects of actions are non-deterministic can be modeled as Markov decision processes. Planning problems are usually goal-directed. This paper proposes several techniques for exploiting the goal-directedness to accelerate value iteration, a standard algorithm for solving Markov decision processes. Empirical studies have shown that the techniques can bring about significant speedups.

Keywords: decision-theoretic planning, Markov decision processes, value iteration, efficiency.

## 1 INTRODUCTION

In a *Markov decision process* (MDP), an agent must, at each time point, choose an action from a finite set $\mathcal{A}$ of possible actions and execute the action. Executing an action $a$ has two consequences: The agent receives an immediate reward $r(s,a)$, which depends on the current state $s$ of the world as well as the action executed, and the world probabilistically moves into another state $s'$ according to a *transition probability* $P(s'|s,a)$.

The action is chosen based on the current state of the world. A *policy* $\pi$ prescribes an action for each possible state. In other words, it is a mapping from the set $\mathcal{S}$ of all possible states to $\mathcal{A}$. The set of possible states is assumed to be finite in this paper. The quality of a policy $\pi$ is measured by its *value function* $V^\pi(s)$; for any state $s$, $V^\pi(s)$ is the expected total discounted reward $V^\pi(s)$ the agent, under the guidance of $\pi$, receives starting from an initial state $s$. A policy $\pi^*$ is *optimal* if $V^{\pi^*}(s) \geq V^\pi(s)$ for any state $s$ and any other policy $\pi$. The value function of an optimal policy is usually referred to as the *optimal value function* and denoted by $V^*$.

MDPs have been studied extensively in the dynamic programming literature (e.g. Howard 1960, Puterman 1990, Bertsekas 1987, White 1993). Dean and Kanazawa (1989) and Dean and Wellman (1991) initiated the use of MDPs in planning problems where effects of actions are not deterministic. Planning problems typically have a large number of states. Solving MDPs with large state space has hence become a hot topic in AI (e.g. Dean *et al* 1993, Boutillier *et al* 1995).

A planning problem can be modeled as an MDP in such way that (1) there is a state designated to be the goal and an action called declare-goal; (2) the reward function $r(s,a)$ is given by

$$r(s,a) = \begin{cases} 1 & \text{if } a\text{=delcare-goal and } s\text{=goal}, \\ 0 & \text{otherwise}; \end{cases} \quad (1)$$

and (3) the action declare_goal cannot be executed more than once. We call MDPs with such properties *goal-directed MDPs*.

Value iteration is a standard algorithm for solving MDPs. This paper proposes several techniques for accelerating value iteration in goal-directed MDPs. Let us begin with a brief review of value iteration and of previous works on speeding up value iteration.

## 2 VALUE ITERATION

A *value function* is a mapping from the set $\mathcal{S}$ of possible states to the real line. Given a value function $V$, define another value function $TV$ by

$$TV(s) = max_a[r(s,a) + \gamma \sum_{s'} P(s'|s,a)V(s')] \quad (2)$$

for each state $s$, where $0 \leq \gamma < 1$ is a discount factor. $T$ is an mapping from the space of value functions to itself. For any function $V$, its *norm* $||V||$ is defined by $||V|| = max_s |V(s)|$. $T$ is the contraction mapping (e.g. Puterman 1990) in the sense that for any two value functions $U$ and $V$,

$$||TU - TV|| \leq \gamma ||U - V||.$$

For any positive number $\epsilon$, we say that a value function $V$ is $\epsilon$-*contracted* if

$$||V - TV|| \leq \epsilon.$$

The optimal value function satisfies the *optimal equation*

$$V^* = TV^*,$$

and hence is 0-contracted.

A value function $V$ induces a policy through

$$\pi(s) = arg\ max_a [r(s,a) + \gamma \sum_{s'} P(s'|s,a) V(s'))]. \quad (3)$$

If the value function is $\epsilon$-contracted for a small number $\epsilon$, the induced policy is "good enough" in the sense that

$$||V^\pi - V^*|| \leq \frac{2\epsilon\gamma}{1-\gamma}. \quad (4)$$

Proof of this inequality can be found in, for instance, Puterman (1990). It is evident the policy induced by the optimal value function is optimal.

Value iteration (VI) (Bellman 1957) starts with an arbitrary value function and improves it iteratively until the value function becomes $\epsilon$-contracted. Here is the pseudo-code.

```
VI
  1. Choose an initial value function V_0 and
     set n=0.
  2. V_{n+1} = TV_n.
  3. If ||V_{n+1} - V_n|| > ε, increment n by 1 and
     go to step 2.
  4. Else return V_{n+1}.
```

Since $T$ is a contraction mapping, $||TV_{n+1} - V_{n+1}|| = ||TV_{n+1} - TV_n|| \leq \gamma ||V_{n+1} - V_n|| \leq \epsilon$. Hence, $V_{n+1}$ is $\epsilon$-contracted.

## 3   PREVIOUS WORK

VI converges geometrically at rate $\gamma$. Convergence is slow when $\gamma$ is close to 1. Various modifications to standard VI have been proposed and all have been theoretically or empirically shown to lead to faster convergence. Morton and Wecker (1977) suggest that one, before applying the operator $T$ in step 2, subtracts an appropriate value function from $V_n$ and MacQueen (1969) proposes to subtract $V_n(s_0)$ — the value of $V_n$ itself at a predetermined state $s_0$. The aggregation/disaggregation techniques introduced by Schweitzer *et al* (1985) and Bertsekas and Castanon (1989) interleave standard VI steps with aggregation/disaggregation steps, which improve the current value function by solving the optimality equation for an simpler MDP obtained from the original MDP through state aggregation. Dean and Lin (1995) and Dean *et al* (1997) decompose an MDP with a large state space into a number of MDPs with smaller state spaces through state aggregation. The smaller MDPs are solved using standard VI and their solutions are used to construct a solution to the original MDP.

Three pieces of previous workd are of direct relevance to this paper. The first one is the Gauss-Seidel variant of standard VI proposed by Hastings (1969). Let $\rho$ be an ordering among the possible states. Instead of the operator $T$ defined in equation (2), the Gauss-Seidel variant uses another operator $T'$ to improve the current value function. For any value function $V$, $T'V(s)$ is defined for each state $s$ by starting from the state that comes first in the ordering $\rho$ and moving backwards. The values $T'V(s)$ for earlier states are used in defining the values for later states. Specifically, $T'$ is given by

$$T'V(s) = max_a [r(s,a) + \gamma \sum_{s'} P(s'|s,a) \hat{V}(s'),] \quad (5)$$

where $\hat{V}(s') = T'V(s')$ when $s'$ comes before $s$ in the ordering $\rho$ and $\hat{V}(s') = V(s')$ otherwise.

The anytime algorithm presented in Dean *et al* (1993) is also closely related to the methods to be proposed in this paper. The algorithm restricts standard VI inside an envelope, a subset of possible states, that contains at least one path from the initial state to the goal state. The envelope is gradually enlarged to get better and better solutions.

Boyan and Moore (1996) study value iteration in acyclic goal-directed MDPs. A goal-directed MDP is *acyclic* if once leaving a state, the world can never come back to that state again. Boyan and Moore point out that value iteration for goal-directed acyclic MDPs can be carried out in one sweep by starting from the goal state and working backwards. Thereby the amount of computations needed to compute the optimal value function is reduced to that needed in one iteration of standard VI. The method is an extension of the DAG-SHORTEST-PATH algorithm (Cormen *et al* 1990) for finding shortest path in acyclic graphs.

## 4   PARSIMONIOUS VALUE ITERATION

We introduce several new variants to standard VI for goal-directed MDPs. Called *parsimonious value iteration* (PVI), the first variant relies on the following intuition. Suppose value iteration begins with the zero value function. Then at early iterations, the value function remains zero for states far away from the goal. At later iterations, the value function does not change much for states close to the goal. The number of states whose values change significantly from one iteration to the next can be much smaller than the total number of states. At each iteration, PVI updates the value for a state only when the value is expected to change significantly.

Specifically, PVI begins with the zero value function. At each iteration $n+1$ ($n \geq 1$), PVI performs a test to detect states whose values change substantially from iteration $n$ to iteration $n+1$. The value of a state is updated only if it passes the test.

Let $V_{n-1}$ and $V_n$ be the value functions PVI computed at the previous two iterations. At the current iteration $n+1$, PVI does not update the value for a state $s$ if $|V_n(s') - V_{n-1}(s')| \leq \delta$ for a small positive constant $\delta$ and each state $s'$ such that $max_a P(s'|s,a) > 0$. Since the number of states reachable from $s$ by executing one action is usually small, this test is cheap. It is usually much cheaper than calculating $TV_n(s)$, especially when one maintains a list of nodes reachable from each state by executing one action.

Theoretical underpinings of the test are as follows. If the value functions $V_n$ and $V_{n-1}$ were the value functions computed by VI, one could easily show that if $s$ passes the test then $|V_{n+1}(s) - V_n(s)| \leq \gamma \delta$. In other words, the value for $s$ does not change much from iteration $n$ to iteration $n+1$.

Here is the pseudo-code for PVI.

PVI

1. $V_0(s) = 0$ for any $s$, $n = 0$.
2. **For** each state $s$,
   (a) If $n \geq 1$ and $|V_n(s') - V_{n-1}(s')| \leq \delta$ for all $s'$ such that $max_a P(s'|s,a) > 0$,
   $$V_{n+1}(s) = V_n(s).$$
   (b) **Else**
   $$V_{n+1}(s) = TV_n(s).$$
3. If $||V_{n+1} - V_n|| > \epsilon$, increment $n$ by 1 and go to step 2.
4. **Else** return $V_{n+1}$.

There is no guarantee that the value function returned by PVI is $\epsilon$-contracted. However, the value function should be close to be $\epsilon$-contracted. We suggest to use PVI as a preprocessing step to VI, i.e. to use the value function it returns as the initial value function of VI. This way an $\epsilon$-contracted value function can be obtained. Since the value function return by PVI is close to be $\epsilon$-contracted, VI should terminate in a small numer of steps. In our experiments, it terminated in just one iteration.

The idea behind PVI is rather similar to the idea underlying Boyan and Moore's one-sweep algorithm; start from the goal and work backwards. PVI does not assume acyclicity and hence is more general. When the MDP is acyclic, it is almost identical to the one-sweep algorithm.

PVI is also related to the anytime algorithm by Dean *et al* (1993) in the sense that values are updated only for some states at each iteration. The difference lies in the fact that in PVI the states whose values are updated change from iteration to iteration, while in the anytime algorithm whether the value for a state is updated depends on whether it is in the envelop and does not change with iteration. Also the entire value iteration process needs to be carried out for each envelop.

## 5   GREEDY AND DOUBLE VALUE ITERATION

Even though the test in PVI is cheap, the fact that it has to be carried out for each state is somewhat unsatisfying. Greedy value iteration (GVI) avoids the test by working in a way similar to DAG-SHORTEST-PATH.

Before describing GVI, we need to introduce the concept of ideal reachability. We say a state $s'$ is *ideally reachable in one step* from another state $s$ if after executing a certain action in state $s$ the probability of the world ending up in state $s'$ is the highest. A state $s_k$ is *ideally reachable in k steps* from another state $s_0$ if there are states $s_1, \ldots, s_{k-1}$ such that $s_{i+1}$ is ideally reachable from $s_i$ in one step for all $0 \leq i \leq k-1$. Any state is ideally reachable from itself in 0 step.

For any state $s$, let $d(s)$ be the minimum number of steps in which the goal is ideally reachable from $s$. We shall refer to $d(s)$ as the *distance* from $s$ to the goal.

At each iteration $n$, GVI only updates the values for the states from which the goal is ideally reachable in $n$ steps. Let $N$ be the maximum number of steps that the goal can be ideally reached from any state. Then GVI terminates in exactly $N$ iterations. For later con-venience, we assume that GVI takes a value function as input and uses it as the initial value function. Here is the psuedo-code for GVI.

GVI($V_0$)

1. **For** $n=0$ to $N$,
    - **For** each state $s$,
    (a) **If** $d(s)=n$, set
    $$V_{n+1}(s) = TV_n(s).$$
    (b) **Else**
    $$V_{n+1}(s) = V_n(s).$$
2. **Return** $V_N$.

When the MDP is acyclic, GVI is identical to Boyan and Moore's one-sweep algorithm and hence returns the optimal value function. When the MDP is cyclic, however, the value function it returns could be of very poor quality. Using it as a preprocessing step to VI might not help much.

On the positive side, the amount of computations GVI does is identical to that carried out by one iteration of standard VI. Also because GVI is an approximation of the entire value iteration process, the extent to which it improves the input value function should be greater than that brought about by one iteration of standard VI. Thus we can expect VI to converge faster if the second line is replaced by "$V_{n+1} = \text{GVI}(V_n)$". This leads to new algorithm called double value iteration (DVI).

DVI
1. Choose an initial value function $V_0$ and set $n=0$.
2. $V_{n+1} = \text{GVI}(V_n)$.
3. **If** $||V_{n+1}(s) - V_n(s)|| > \epsilon$, increment $n$ by 1 and go to step 2.
4. **Else** return $V_n$.

As it turns out, DVI can be described directly without the reference to GVI. At each iteration, it uses a new operator $T'$, instead of the operator $T$ given in Equation (2), to update the value function $V_n$. For any value function $V$, $T'V(s)$ is defined for each state $s$ by starting with the goal state and gradually moving away. The value $T'V(s)$ for a state $s$ is defined after the values $T'V(s')$ for all the states $s'$ closer to the goal than $s$ having been defined. It is given by

$$T'V(s) = max_s[r(s,a) + \gamma \sum_{s'} P(s'|s,a)\hat{V}(s',s),]$$

where

$$\hat{V}(s',s) = \begin{cases} TV(s') & \text{if } d(s') < d(s), \\ V(s') & \text{otherwise.} \end{cases}$$

It can be proved that $T'$ is also a contract mapping. Hence the value function returned by DVI is $\epsilon$-contracted.

It is evident see that DVI is almost identical to the Gauss-Seidel variant of standard VI, except that it proposes one particular way to order the possible states; the states are ordered according to their distances to the goal. By introducing DVI through GVI, we hope to provide another way of looking at the Gauss-Seidel variant of standard VI in the context of goal-directed MDPs.

## 6   IMPROVING PVI

The alternative understanding of DVI can be used to improve PVI. We call the improved algorithm PVI1. The pseudo-code is as follows.

PVI1

1. $V_0(s)=0$ for any $s$, $n=0$.
2. **For** $m=0$ to $N$,
    (a) **For** each state $s$ such that $d(s)=m$
    (b) **If** $n \geq 1$ and $|V_n(s') - V_{n-1}(s')| \leq \delta$ for all $s'$ such that $max_a P(s'|s,a) > 0$,
    $$V_{n+1}(s) = V_n(s).$$
    (c) **Else**
    $$V_{n+1}(s) = T'V_n(s).$$
3. **If** $||V_{n+1} - V_n|| > \epsilon$, increment $n$ by 1 and go to step 2.
4. **Else** return $V_{n+1}$.

As PVI, PVI1 should be used as a preprocessing step to VI.

## 7   EXPERIMENTS

Preliminary experiments have been carried to compare the algorithms proposed in this paper with standard value iteration. Four office environment navigation problems borrowed from Cassandra *et al* (1996) were used. The problems differ in corridor layout and the total number of states. There are two sets of transition probabilities, referred to as standard and noisy transition probabilities respectively. Effects of actions are less certain under noisy transition probabilities than under standard transition probabilities.

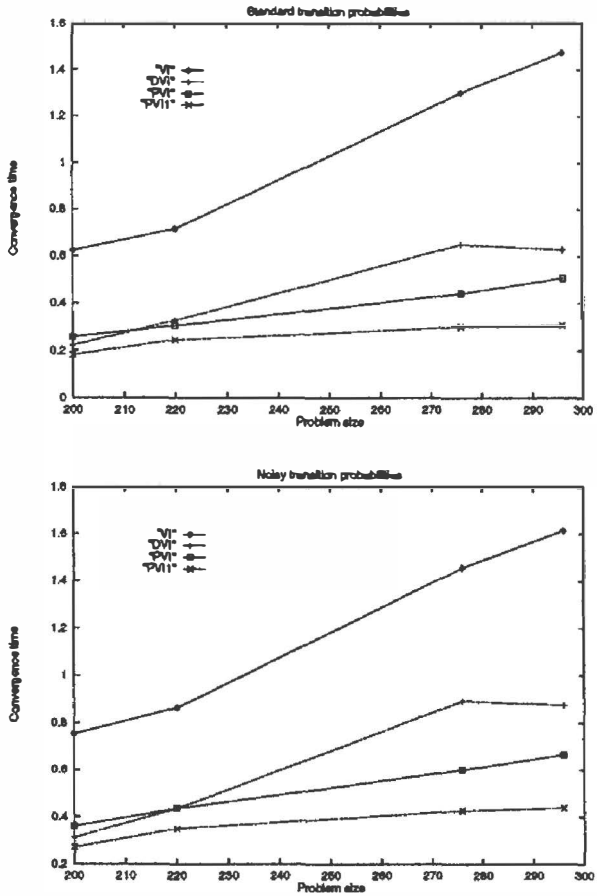

Figure 1: Convergence times of the algorithms in four navigation problems.

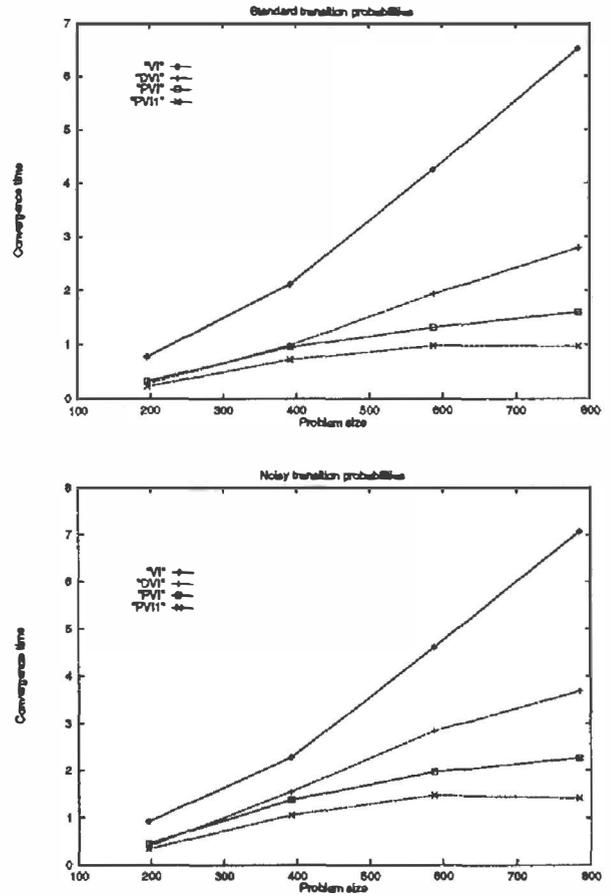

Figure 2: Differences in performance among the algorithms as problem size increases.

The threshold for the Bellman residual was set at 0.001 and the discount factor at 0.99. Figure 1 shows that convergence times of the algorithms in the four problems. The X-axis represents the sizes of the problem, while the Y-axis represents convergence time in CPU seconds. Data were collected using a SPARC20. The curves VI and DVI display the convergence times of VI and DVI respectively, while PVI and PVI1 display the convergence times for the combinations of PVI and PVI1 with VI.

Under both standard and noisy transition probabilities, DVI and PVI converges much faster than VI and PVI1 converges even faster. DVI converges slightly faster than PVI in the smallest problem but slower in all other problems. Performances of all algorithms are slightly worse under noisy transition probabilities than under standard transition probabilities. Their differences are also slightly larger.

To gain an idea about how the comparisons change with problem sizes, we made copies of one environment and glue them together to form larger environments.

The convergence times are shown in Figure 2. We see that the differences in performance among the algorithms become larger as the problem size increases. In the smallest problem PVI1 converges about three times faster than VI, while in the largest problem it converges six times faster.

## 8  CONCLUSIONS AND FUTURE DIRECTIONS

We propose several techniques for exploiting the goal-directedness of planning problems to speed up value iteration for their MDP models. Empirical studies have shown that the techniques can bring about significant speedups.

MDPs assume perfect observation of the state of the world. In many real-world problems, one does not know the true state of the world. Such problems can be modeled as partially observable MDPs (POMDPs). POMDPs are much harder to solve than MDPs. We are currently investigating the possibility of applying the ideas introduced in this paper to POMDPs.

### Acknowledgements

We thank Peter Dayan, Thomas L. Dean, and Michael Littman for pointers to references and thank Wenju Liu and D. Y. Yeung for useful discussions. Research was supported by Hong Kong Research Council under grants HKUST 658/95E and Hong Kong University of Science and Technology under grant DAG96/97.EG01(RI).

## References

[1] R. Bellman (1957), *Dynamic Programming*, Princeton University Press.

[2] D. P. Bertsekas and D. A. Castanon( 1989 ), Adaptive Aggregation for Infinite Horizon Dynamic Programming, *IEEE trans. on auto. control*, vol 34, No 6, 1989.

[3] D. P. Bertsekas (1987), *Dynamic Programming: Deterministic and Stochastic Models*, Prentice-Hall.

[4] C. Boutillier, R. Dearden and M. Goldszmidt (1995), Exploiting Structures In Policy Construction, In *Proceedings of IJCAI'95*. pp. 1104-1111.

[5] J. A. Boyan and A. W. Moore (1996), Learning Evaluation Functions for Large Acyclic Domains." In L. Saitta(ed.), *Machine Learning: Proceedings of the Thirteenth International Conference*, Morgan Kaufmann.

[6] T. H. Cormen, C. E. Leiserson, and R. L. Rivest (1990), *Introduction to Algorithms*, MIT Press.

[7] T. L. Dean and K. Kanazawa (1989), A Model for Reasoning about Persistence and Causation, *Computational Intelligence*, 5(3), pp. 142-150.

[8] T. L. Dean, R. Givan, and S. Leach, Model reduction techniques for computing approximately optimal solution for Markov decision processes, in *Proceedings of the Thirteenth Conference on Uncertainty in Artificial Intelligence*.

[9] T. L. Dean, L. P. Kaelbling, J. Kirman, and A. Nicholson (1993), Planning with Deadlines in Stochastic Domains, In *Proceedings of the Eleventh National Conference on Artificial Intelligence*, Washington, DC.

[10] T. L. Dean and S. H. Lin (1995), Decomposition techniques for planning in stochastic domains, TR CS-95-10, Department of Computer Science, Brown University, Providence, Rhode Island 02912, USA.

[11] T. L. Dean and M. P. Wellman (1991), *Planning and Control*, Morgan Kaufmann.

[12] N. A. J. Hastings (1969), Optimization of Discounted Markov Decision Problems, *Oper. Res. Quart.*, 20, 499-500.

[13] R. A. Howard (1960), Dynamic Programming and Markov Decision Processes, Wiley, London.

[14] J. MacQueen (1969), A Modified Dynamic Programming Method for Markov Decision Problems, *J. Math. Anal. Appl.*, 14, 38-43.

[15] T. E. Morton and W. E. Wrecker (1977). Decision Ergodicity and Convergence for Markov Decision Processes. *Management Sci.*, 23, 890-900.

[16] M. L. Puterman (1990), Markov Decision Processes, in D. P. Heyman and M. J. Sobel (eds.), *Handbooks in OR & MS*, Vol. 2, pp. 331-434, Elsevier Science Publishers.

[17] P. J. Schweitzer, M. Puterman, and K. W. Kindle, Iterative aggregation-disaggregation procedures for solving discounted semi-Markovian reward processes, *Operations Research*, 33, pp. 589-606, 1985.

[18] D. J. White (1993), *Markov Decision Processes*, John Wiley & Sons.